\documentclass[conference]{IEEEtran}
\IEEEoverridecommandlockouts
\usepackage{cite}
\usepackage{amsmath, amssymb, amsfonts}
\usepackage{algorithmic}
\usepackage{graphicx}
\usepackage{textcomp}
\usepackage{xcolor}
\def\BibTeX{{\rm B\kern-. 05em{\sc i\kern-. 025em b}\kern-. 08em
    T\kern-. 1667em\lower. 7ex\hbox{E}\kern-. 125emX}}
\begin{document}

\title{Robust Two-Stream Multi-Feature Network for Driver Drowsiness Detection \\}

\author{\IEEEauthorblockN{Qi Shen$^{\star \dagger}$, Shengjie Zhao$^{\star \dagger}$, Rongq Zhang, Bin Zhang$^{\star}$}
	\IEEEauthorblockA{$^{\star}$School of Software, Tongji University, Shanghai, China\\
		$^{\dagger}$Key Laboratory of Embedded System and Service Computing, Ministry of Education, China\\
		Email: \{1833029, shengjiezhao, rongqingz, duckie\}@tongji.edu.cn}
}

\maketitle

\begin{abstract}
Drowsiness driving is a major cause of traffic accidents and thus numerous previous researches have focused on driver drowsiness detection. Many drive relevant factors have been taken into consideration for fatigue detection and can lead to high precision, but there are still several serious constraints, such as most existing models are environmentally susceptible. In this paper, fatigue detection is considered as temporal action detection problem instead of image classification. The proposed detection system can be divided into four parts: (1) Localize the key patches of the detected driver picture which are critical for fatigue detection and calculate the corresponding optical flow. (2) Contrast Limited Adaptive Histogram Equalization (CLAHE) is used in our system to reduce the impact of different light conditions. (3) Three individual two-stream networks combined with attention mechanism are designed for each feature to extract temporal information. (4) The outputs of the three sub-networks will be concatenated and sent to the fully-connected network, which judges the status of the driver. The drowsiness detection system is trained and evaluated on the famous Nation Tsing Hua University Driver Drowsiness Detection (NTHU-DDD) dataset and we obtain an accuracy of 94.46\%, which outperforms most existing fatigue detection models. 
\end{abstract}


\section{Introduction}
	Over the last decades, driver drowsiness is one of the main causes of traffic accidents. About 20\% to 30\% of the crash can be owing to fatigue driving. Accordingly, many researches have been conducted to solve this problem by detecting the drivers' fatigue effectively and sending out alerts timely.  Relevant studies have pointed out several features for discovering the sleepiness of the driver. The drowsiness detection models can be generally classified into three categories, that is vehicle-based models, physiology-based models, and face-based ones. Real-time vehicle parameters are readily available but the challenge lies in that the correlative feature about drivers is hard to extract and it is often too late when the detection system finds any exception. Existing researches have demonstrated that physiology-based methods usually have a better detection accuracy, because fatigue changes drivers' physiological condition rapidly, e.g., electroencephalogram (EEG), blood pressure, and heart rate. However, the installation of the apparatus collecting drivers' physiological parameters is both expensive and inconvenient. Furthermore, both vehicle-based and physiology-based techniques are susceptible to external factors, like the weather and the physical condition of the driver. It means that these fatigue detection models may become invalid sometimes. Different from them, face-based techniques make a proper trade off. The driver's facial expression can be caught by cameras easily. In addition, face-based models are accurate and reliable in most circumstances.

\begin{figure}
	\centering
	\includegraphics[width=0.9\linewidth, height=0.32\textheight]{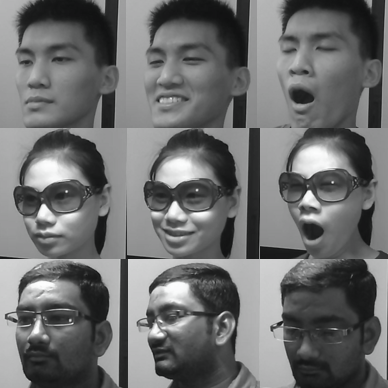}
	\caption{Sample frames from the NTHU-DDD dataset}
	\label{fig:dataset}
\end{figure}

With the rapid development of deep learning, neural networks gradually become the substitute for traditional face-based drowsiness detection methods. Traditional dorwsiness detection networks try to catch some typical behaviors like yawning and head tilt by signal image. All the features involved are theoretically valid and it is not complicated to construct an effective network and obtain high accuracy. But they are incapable to extract subtle sleepiness and the detection system is confused occasionally. For example, talking is often confused with yawning. With this in mind, temporal information is utilized in our system. 
Instead of classic temporal-based models Hidden Markov Model (HMM) and Recurrent Neural Network (RNN), two-stream model is chosen as the base network that simultaneously utilizes the original images and optical flow information. For more precise detection, both subtles features like eyes and remarkable features like yawning are fused in our system. Considering better detection of early drowsiness and high correlation between eyes feature and fatigue, attention  mechanism is applied in eyes feature extraction. It has a certain effect which can be inferred from experiments.

Illumination is a potential factor affecting the detection accuracy of image-related systems, especially for driver drowsiness detection because of  changeable driving environment. In order to design more robust detection system, CLAHE is applied to preprocess the detected pictures. For better comparison, NTHU-DDD is applied in our system and Fig. 1 shows some sample frames from the dataset.

\textbf{}


To summarize, our contributions are listed as follow: 

\begin{itemize}
\item In order to extract features more effectively and reduce the impact of the detection environment, we employ a more efficient pretreatment for the detected data. Specifically, key areas of the picture are cropped instead of using the entire picture for detection. CLAHE is used to balance the exposure of images and makes the details more visible.

\item We propose to fuse multiple features in our system to improve the detection accuracy and the robustness of the model. We design a two-stream-based detection network for each feature. In this manner, temporal information can be effectively mined. Moreover, we also employ the attention mechanism to optimize our eyes detection network. 

\item We evaluate our proposed drowsiness detection system on the NTHU-DDD \cite{b0} dataset, and the results demonstrate that our proposed system can achieve an accuracy of 94.46\% eventually, which exceeds most existing drowsiness detection models. 
\end{itemize}

\section{Related Works}

\subsection{Fatigue Detection Features}
Due to the serious limitations, the proportion of the studies based on the driver's physiological conditions and vehicle driving parameters is not large. Face-based methods are the major direction. In the beginning, people thought about detecting fatigue information through various obvious fatigue characteristics, like eye blink and yawning, but all of them have restrictions in some situations. There is still one issue to consider: how to eliminate or decrease the impact of different light conditions. Gamma Correction is applied in \cite{b1} to enhance image contrast, and it was proved by experiment that the image preprocessing contributes to better results. Therefore, CLAHE is applied in our system to mitigate the influence of illumination. 

Drowsiness eyes detection is a temporal-based problem. Ahmad and Borolie \cite{b2} proposed the eyes detection based drowsiness system. Eyes and head were positioned through the Viola-jones algorithm and the primary contribution for drowsiness detection is eye-blink. Drowsiness alert is sent by the system when the blink rate is below the threshold. Holding an identical view, the fatigue monitoring system proposed by Rahman \emph{et al}\cite{b3} has a similar procedure. Extract eye-blink feature and detect drowsiness, but particular eye-blink detection methods are applied in the system. Upper corner points and lower corner points of the eyelid are detected by Harris corner detector. Then the upper mid-point is calculated by two upper corner points and the lower mid-point is calculated by two lower corner points. Eventually, the distance between upper mid-point and lower mid-point provides the status of eyes and decides the status of drivers. Nevertheless, the accuracy of these models will be greatly reduced when the driver wears sunglasses. 

It's not a difficult problem to detect yawn and a high recall rate is always acquired. An efficient yawning detection system was proposed by Alioua \emph{et al}. \cite{b4}. It focused on locating the mouth and calculated the degree of mouth opening. How to distinguish between laughter, normal talk, and yawn become a rough spot when the fatigue alert system only based on yawn detection. Those features are pretty similar sometimes. It seems that the detection of yawn is easily disturbed if the system judges drowsiness by one frame image, so temporal-based methods are used in our system when detecting yawn.

Due to the limitations of single feature-based models, some researchers make use of the entire facial image. Jie Lyu\cite{b5} proposed a robust driver drowsiness detection model MCNN. Besides eyes, nose, and mouth, original face image and several facial patches are put into the detection model. Local and global information are fully utilized. And RNN with multiple LSTM blocks were applied to dig temporal features. They achieved 90.05\% accuracy on NTHU-DDD dataset finally. However, in order to extract drowsy information, some redundant features are undoubtedly involved and there is still some space for optimization. Instead of taking advantage of the whole image directly, most of the fatigue detection system combined various features. \cite{b6} detected eye status and yawning simultaneously. Eye closure and head position were detected for the driver's drowsiness detection in \cite{b7}. Although physiological parameter-based techniques and vehicular condition-based techniques are defective, it is indeed valuable when they are set to auxiliary features.  \cite{b8} proposed the hybrid approaches of drowsiness detection. Heart rate, vehicle velocity, and eyelid closure are caught in the system for monitoring the abnormal status of the driver. \cite{b9} detects drowsiness through head movement and heart rate obtained by frame difference algorithm and R-peak detection algorithm. In addition to the accuracy improvement, there is no doubt that mixing diverse signals makes the model more robust. 
\begin{figure*}
	\centering
	\includegraphics[width=0.85\linewidth, height=0.35\textheight]{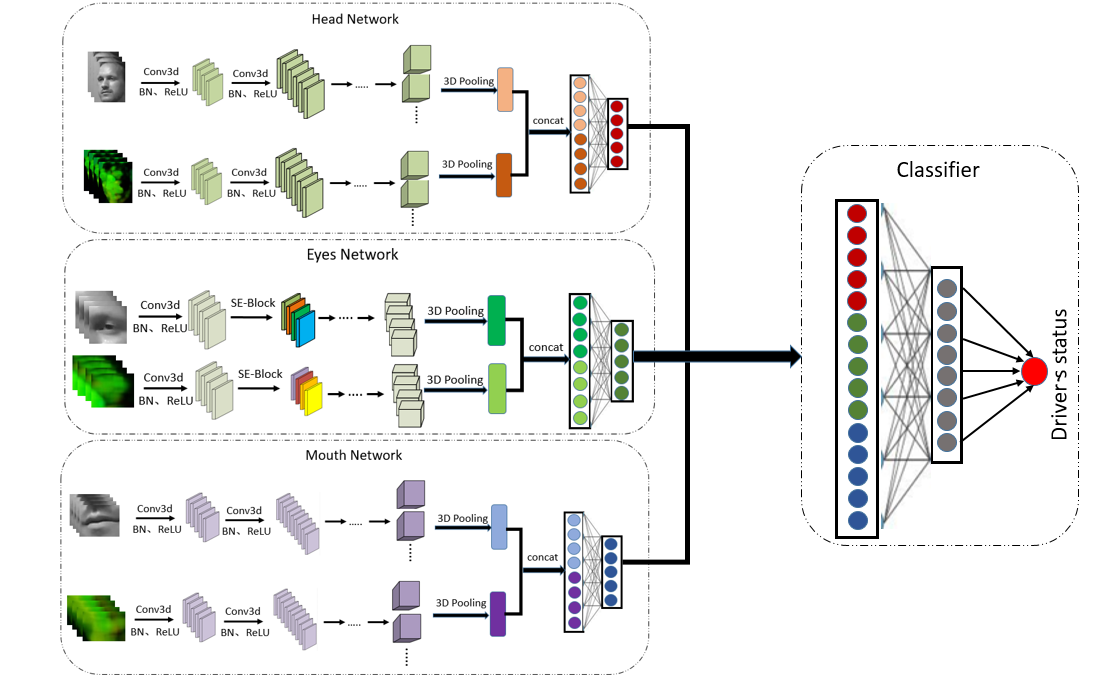}
	\caption{Proposed multi-feature model for fatigue detection}
	\label{fig:network1}
\end{figure*}

\subsection{Fatigue Detection Technology}

Almost the earliest drowsiness detection models concentrated on mathematical techniques, like bio-mathematical models which predict drowsiness by prior information such as duration of sleep, duration of wakefulness, and sleep history. In \cite{b10} they proposed Two Process Model to predict fatigue levels. The Three Process Model \cite{b11} further exploited the duration of sleep and wakefulness to get more accurate fatigue information. 

As a traditional machine learning classification method, SVM was wildly utilized in drowsiness detection models.  \cite{b12,b13,b14} assessed drowsiness levels by eye status and SVM classifier. Taking temporal information into account, HMM played a crucial role in \cite{b15,b16,b17}. With the rapid development of deep learning, SVM is replaced by Convolution Neural Network (CNN) and HMM is replaced by LSTM. Although better results were obtained, it did increase the amount of calculation.

\section{Proposed Methods}

In this paper, various facial features are extracted and fuse in our drowsiness detection model. CLAHE is used for diverse light conditions. Two-stream based sub-networks fully make use of temporal information. The eyes detection network is optimized according to the characteristic of the eyes features. The extracted features are concatenated for the final classification. We train and test on NTHU-DDD dataset and achieve the detection accuracy of 94.46\% eventually.

\subsection{Key Patches and Optical Flow}
Multiple facial drowsiness features are fused in our detection model for more robust and precise detection. For effective extraction, we localize critical face areas first. Considering both complexity and precision, Multi-task Convolutional Neural Network (MTCNN) \cite{b18} is selected for the acquirement of mouth, eyes, and head patches.

As a classic and high-performance face detection model, MTCNN is composed of three lightweight cascaded convolution networks P-Net, R-Net, and O-net. The three networks take in the image and calibrate the face bounding boxes and key points step by step. The bounding boxes and alignment of the face are generated by P-Net and corrected by R-Net and O-Net. Proposal Network (P-Net) implements coarse localization. To remove the candidate boxes with lower scores, Non-Maximum Suppression (NMS) is utilized to further decrease the number of the candidate boxes. The scores of the bounding boxes determined as follows:

$$
s_i =  \begin{cases}
s_i & \text{IoU(M,}b_i\text{)}<N\\
0 & \text{IoU(M,}b_i\text{)} \geq N
\end{cases} \eqno (1)    
$$

\begin{figure}
	\centering
	\includegraphics[width=1.0\linewidth]{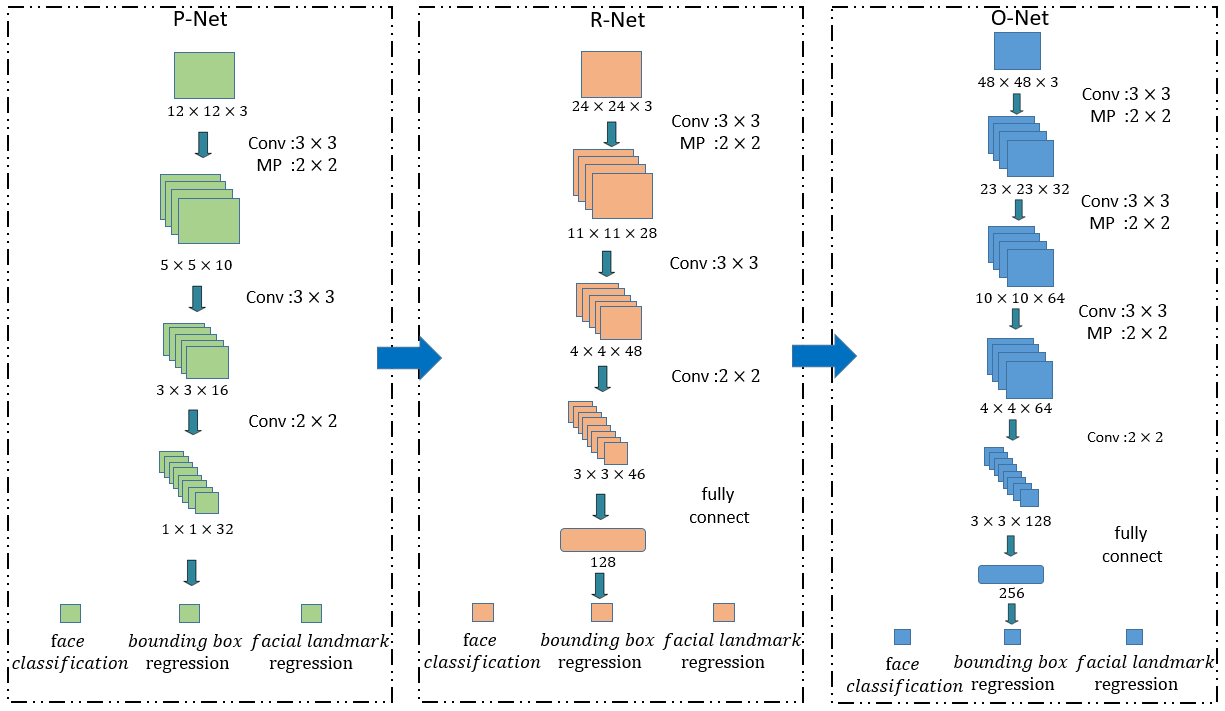}
	\caption{The structure of MTCNN}
	\label{fig:mtcnnn}
\end{figure}

For calibration, Refine Networks (R-Net) takes the result of the first stage as input and rectify the bounding boxes by regression. Similar to P-Net, Output Network (O-Net) increases the location precision based on the result of P-Net and NMS. O-Net produces the final bounding boxes and face alignment points. According to the bounding boxes and face landmarks, the required key parts are cropped from the original image. 

\subsection{Contrast Limited Adaptive Histogram  Equalization}
Minimizing the impact of various lighting conditions is a non-trivial problem in computer vision. The precision of the face-based models generally decrease if the illumination of the driving environment changes. Because of the changing driving environment, it is more challenging for video-based fatigue detection models. There may be strong and uneven sunlight in the car, or driving at night sometimes. Some proper measures must be adopted, and the most effective solution is image preprocessing. 

For overexposure and underexposure images, histogram equalization algorithm is a common treatment for rectifying the image's contrast. Image histogram represents the pixel intensity distribution of a digital image. In numerous image-related fields, CLAHE is applied for abnormal exposure pictures and it stretches the distribution of pixel intensity and enhances the local contrast of the image. In other words, the pixel values are redistributed. The pixel distribution of the original image is mapping to another value. To achieve histogram equalization, Cumulative Distribution Function (CDF) is required as the mapping function. Except CLAHE, gamma correction and Laplace transform also have a similar effect. 

Traditional histogram equalization algorithms apply the same histogram transformation to each pixel. This strategy is effective for the image with uniform pixel distribution. But for those that contain significant bright or dark areas, they are not able to produce a fantastic effect and Adaptive Histogram Equalization (AHE) algorithm solves the problem. It performs histogram enhancement on each pixel by calculating the transformation function through the neighborhood pixels. CLAHE is a histogram equalization algorithm based on AHE, which overcomes the drawback of excessive amplification of the noise by limiting the contrast. For acceleration, CLAHE adopts a special interpolation algorithm. The detection picture is divided into multiple blocks. The formulation is as follows:

	$$
	\begin{aligned}
	f(D) = & (1-\Delta y)((1-\Delta x)f_{ul}(D)+\Delta x f_{bl}(D))\\
	& +\Delta y((1-\Delta x)f_{ur}(D)+\Delta x f_{br}(D))\\
	\end{aligned} \eqno (2) 
	$$

For each pixel, the mapping values of the four adjacent parts histogram CDF to the pixel are required.   $ \Delta x $ and $ \Delta y $ indicate the distance between the pixel and the center of the left upper block.

For the illumination in the car is sometimes uneven, CLAHE is more suitable for the driver fatigue detection system than traditional histogram equalization algorithms. For the implementation of CLAHE, Opencv library is adopted. Fig. 4 shows the comparison between the normal images and the images processed by CLAHE. It's obvious that the face features, especially the eye features are much more visible when people wear sunglasses. Although the final result has been improved, CLAHE is extremely time-consuming and needs further optimization.

\begin{figure}
	\centering
	\includegraphics[width=0.85\linewidth, height=0.32\textheight]{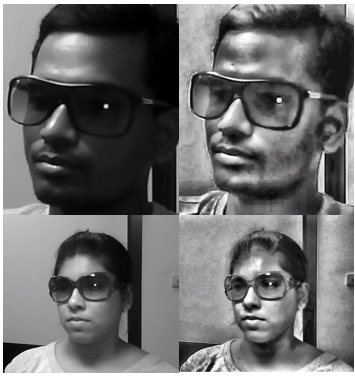}
	\caption{The picture before and after processed by CLAHE}
	\label{fig:clahe}
\end{figure}

\subsection{Two-Steam Detection Network}
In real-world detection situations, it is inevitably problematic if only single frame data are utilized for fatigue detection. Temporal feature-based detection is a more appropriate choice. The fatigue detection based on facial information and temporal features is actually a temporal action detection model, which determines whether the driver is in the state of fatigue based on the facial expression during a certain period of time. 

Two-stream detection network is a classic model in temporal action detection. A series of continuous frames data and corresponding optical flow information are separately sent into network and features are merged eventually. Original images would provide spatial information while optical flow information provides motion information. Temporal information is extracted by 3D convolution in the two-stream network. Different from 2D convolution, 3D convolution layers add the time channel, and equation (3) shows the difference between them. Finally, spatial features and motion features are fused to make the judgment of drowsiness. Only a single frame and relevant optical flow are involved in \cite{b1}, and we consider the temporal information is insufficient to detect drowsiness precisely. 
	$$
	\begin{aligned}
		conv_{2d}(i,j)&=\sum\limits_{m,n} x(i-m,j-n)w(m,n) \\
		conv_{3d}(i,j,t)&=\sum\limits_{m,n,k}  x(i-m,j-n,t-k)w(m,n,k) \\
	\end{aligned} \eqno (4) 
	$$

Eye, mouth, and head posture are three major features for face-based fatigued detection and three customized sub-networks are designed for these fatigue features. As shown in Fig. 2, the proposed fatigue detection model consists of three base detection networks. For each sub-network, original images and corresponding optical flow images are processed by several layers of 3D convolution and concatenated after 3D pooling. For the head network, the input patches are resized to $ 224 \times 224 $. Eye and mouth patches are resized to $ 112 \times 112 $. The features extracted by the three networks and fully-connected layers, which produce the judgment of fatigue. To prevent overfitting, L2 regularization is added to convolution layers and dropout is added to fully connected layers. Three sub-networks are pretrained on NTHU-DDD dataset and experiments show that the sub-network pretraining can greatly improve the detection accuracy. We use the cross-entropy loss to train the detection network, which is formulated as:

$$
Loss_{det} = \sum_i -(p_i^{} \cdot \log q_i)  \eqno (5)
$$
\begin{figure}
	\centering
	\includegraphics[width=0.7\linewidth]{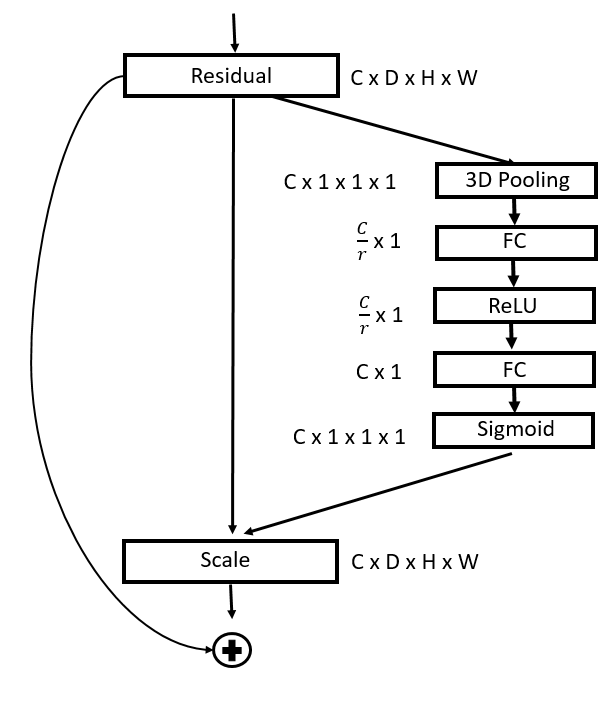}
	\caption{SE-Block}
	\label{fig:se-block}
\end{figure}

\begin{figure}
	\centering
	\includegraphics[width=1.0\linewidth]{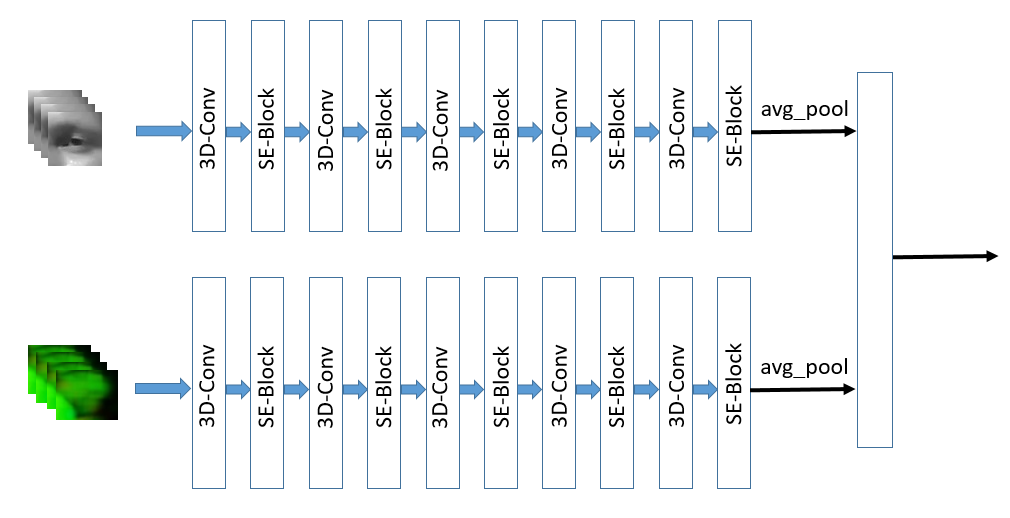}
	\caption{Eyes Network}
	\label{fig:eyenet}
\end{figure}

\subsection{Sequeze and Excitation}
In terms of the three relative drowsiness features, the correlation between eye information and fatigue is the highest. However, it is a challenge to detect drowsiness eyes accurately especially in early fatigue. For this reason, some optimization is applied to the eyes network. Squeeze-and-Excitation Networks (SE-Net) \cite{b19} was affected by the attention mechanism and exploit the relevance between the filter channels to improve the classification performance. The specific procession contains two steps, Squeeze and Excitation. The squeeze operation learns the relationship between the channels by convolution and the excitation operation then applies the relationship between the channels to the filters of the layer.

Fig. 6 shows the structure of the eyes network. SE-Blocks were also added to the mouth network and the head network, but they were removed finally because of little improvement. \cite{b20} mentioned that downsampling in the shallow layers often reduces the effectiveness of the model, so we remove the downsampling operator in the shallow layers and complete that in the last several layers.

\section{Experiments } 

\subsection{Dataset and Data Preprocession}

National Tsing Hua University-Driver Drowsiness Detection (NTHU-DDD) dataset was created by National Tsing Hua University and contains various typical scenarios that are close to real-world driving conditions, such as wearing sunglasses and night driving. It also has abundant kinds of labels, including eyes, mouth, head, and the judgment of fatigue. NTHU-DDD consists of 18 video sets and 20 evaluation videos. One video set was chosen as the validation set and three for the test set. For the amount of the images input each time, we hold that a short time span is not sufficient to determine whether a driver is in the sleepiness state, especially for the early fatigue. Videos in NTHU-DDD dataset comprises 30 frames per second, and we believe that three seconds is enough to judge whether a driver is fatigued. Consequently, we have two schemes for the video clip. One is catching one image every 10 frames, and input 10 pictures each time, the other is catching one image every 3 frames, and input 30 pictures each time. For the latter schema, more precise detection can be achieved, but it also requires more computing resources. Fig. 7 (e) shows the former detection accuracy of 88.7\% and Fig. 7 (d) is 92.8\% for the latter. 

\begin{figure}
	\centering
	\includegraphics[width=0.9\linewidth]{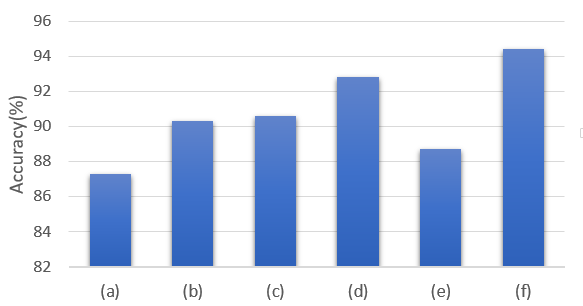}
	\caption{The accuracy of our fatigue detection model base on various configuration}
	\label{fig:res2}
\end{figure}

For the facial features based fatigue detection model, the simplest method is feeding face pictures to the detection network. Fig. 7 (a) shows the result accuracy of fatigue detection by directly processing the continuous sequence of face images and it got an accuracy of 87.3\%. The model in Fig. 7 (b) combines face pictures with optical flow and got an accuracy of 90.3\%. Fig. 7 (d) fuses multi-feature and got an accuracy of 92.8\%. It is obvious that optical flow and multi-feature bring significant performance improvement. More detail about optical flow will be discussed in the next section.

Driver fatigue detection is heavily influenced by driving environments, so the detection system is required to take some special situations into account. For instance, multi-feature fusion is applied in our system in case of the driver wearing sunglasses. For the illumination continues to change from day to night, CLAHE is adopted for image histogram equalization. Fig. 7 (c) shows the result if CLAHE is removed and it got an accuracy of 90.6\%, which indicates that CLAHE indeed contributes to fatigue detection.

\subsection{Critical Feature Extracting}

Two-stream network is the foundation of many famous temporal action detection models, like TSN. 3D convolution draw temporal information through the continuous face images and optical flow information further improve the performance of the network. All facial fatigue features are considered in our model. Table \uppercase\expandafter{\romannumeral1} displays the labels in the NTHU-DDD dataset. Fig. 8 shows the result accuracy of the key area detection models based on two-stream network and only corresponding patches. Fig. 7 (f) shows the experimental results with pre-training and it got an accuracy of 94.46\%. Pretraining benefits the ensemble model to some extent. 

%

\begin{table}[htbp]
	\caption{NTHU-DDD Labels}
	\begin{center}
		\begin{tabular}{|c|c|c|c|}
			\hline
			\textbf{Features}&\multicolumn{3}{|c|}{\textbf{Labels}} \\
			\cline{1-4} 
			\hline
			drowsiness &  Stillness & Drowsy& - \\
			eye & Stillness & Sleepy-eyes & - \\
			mouth & Stillness & Yawning & Talking\&Laughing \\
			head & Stillness & Nodding & Looking aside \\
			\hline
		\end{tabular}
		\label{tab1}
	\end{center}
\end{table}

\begin{figure}
	\centering
	\includegraphics[width=0.9\linewidth]{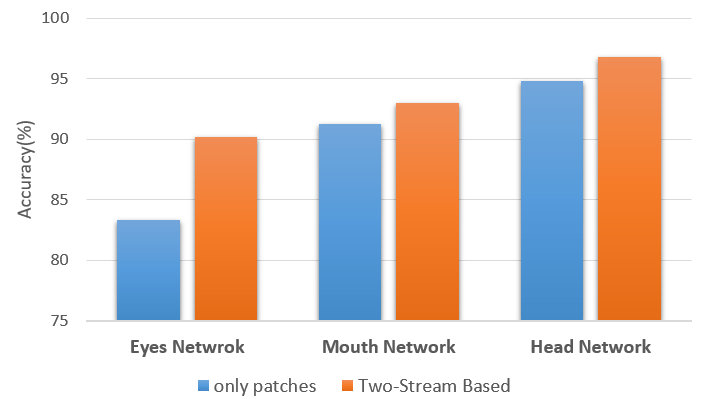}
	\caption{The accuracy of model added and without optical flow}
	\label{fig:res3}
\end{figure}

\subsection{Detection Accuracy}
Table \uppercase\expandafter{\romannumeral2} compares the accuracy between several state-of-the-art drowsiness detection models and ours on NTHU-DDD dataset. The drowsiness detection accuracy of our system exceeds most of the existing models. Instead of concentrating on the depth of the neural network to obtain better results, the width of the model is more attractive to us. In other words, more attention is paid to the number of the feature categories. 

We implement our network using the PyTorch on two Titan X GPU with 12GB memory. To train the sub-network and the final integrated model, we use the Adam optimizer. The initial learning rate is set as $ 1 \times 10^{-4} $ and decreased using the polynomial decay with power of 0.9. 

\begin{table}[htbp]
	\caption{Comparison of Accuracy}
	\setlength{\tabcolsep}{7mm}
	\begin{center}
		\begin{tabular}{c|c|c}
			\hline
			\textbf{Model} & \textbf{\textit{Temporal Features}}& \textbf{\textit{Accuracy}}\\
			\hline
			MSTN \cite{b21} & LSTMs & 85.52\% \\
			\hline
			DDD \cite{b22} & ConvCGRNN& 84.81\%\\
			\hline
			MCNN \cite{b5} & LSTMs & 90.05\% \\
			\hline
			Li \emph{et al}. \cite{b24} & DB-LSTM & 93.6\% \\
			\hline
			ours& Two-Stream &  94.46\% \\
			\hline
		\end{tabular}
		\label{tab1}
	\end{center}
\end{table}
\section{Conclusion and Future work}
To overcome the limitation of existing driver fatigue detection, we propose the multi-features fatigue detection network based on two-stream network. CLAHE is applied to the original picture to reduce the impact of light. To optimize eye information extraction, SE-blocks are added to the eyes network and pretraining is utilized to further improve the performance of the model. Our fatigue detection model achieves an accuracy of 94.46\% on NTHU-DDD dataset, which exceeds most existing fatigue detection models.

Although our system focus on robustness, there are still many conditions that the face feature-based drowsiness detection become invalid, such as when most of the face is covered by hair or hat. Perhaps the detection system should resort to the physiology parameters if facial features are unavailable. To handle these cases, we may attempt to integrate more features in future. 

%





\begin{thebibliography}{00}
\bibitem{b0}	
C. H. Weng, Y. H. Lai, and S. H. Lai, ``Driver drowsiness detection via a hierarchical temporal deep belief network,'' in Asian Conference on Computer Vision, Springer, pp. 117–133, 2016.

\bibitem{b1}
W. Liu, J. Qian, Z. Yao, X. Jiao, and J. Pan, ``Convolutional Two-stream Network Using Multi-Facial Feature Fusion for Driver Fatigue Detection,''  Future Internet, vol. 11, no. 5, 2019.

\bibitem{b2}
R. Ahmad, and J. N. Borole, ``Drowsy driver identification using eye blink
detection,'' International Journal of Innovative Science Engineering and Technology, vol. 6, no. 1, pp. 270–274, Jan. 2015. 

\bibitem{b3}
A. Rahman, M. Sirshar, and A. Khan, ``Real time drowsiness detection
using eye blink monitoring,'' in Proceedings of National Software Engineering Conference, pp. 1–7,  Dec. 2015. 

\bibitem{b4}
Yan \emph{et al.}, ``Video-based classification of driving behavior using a
hierarchical classification system with multiple features,'' International Journal of Pattern Recognition and Artificial Intelligence, vol. 30, no. 5, Art. no. 1650010, 2016. 

\bibitem{b5}
J. Lyu, Z. Yuan, and D. Chen, “Long-term multi-granularity deep framework for driver drowsiness detection,” arXiv preprint arXiv:1801.02325, 2018.

\bibitem{b6}
B. N. Manu, ``Facial features monitoring for real time drowsiness detection,'' in Proceedings of  International Conference on Innovations in Information Technology, pp. 1–4, Nov. 2016. 

\bibitem{b7}
R. Mbouna, S. Kong, and M. G. Chun, ``Visual analysis of eye state and head pose for driver alertness monitoring,''  IEEE Transactions on Intelligent Transportation Systems, vol. 14, no. 3, pp. 1462–1469, Sep. 2013.

\bibitem{b8}
B. G. Lee and W. Y. Chung, ``A smartphone-based driver safety monitoring system using data fusion,'' Sensors, vol. 12, no. 12, pp. 17536–17552, Dec. 2012. 


\bibitem{b9}
M. Awais, N. Badruddin, and M. Drieberg, ``A hybrid approach to detect
driver drowsiness utilizing physiological signals to improve system performance and wearability,'' Sensors, vol. 17, no. 9, p. 1991, Aug. 2017. 

\bibitem{b10}
A. A. Borbély, ``A two process model of sleep regulation,'' Hum Neurobiol., vol. 1, no. 3, pp. 195–204, 1982. 


\bibitem{b11}
T. Åkerstedt and S. Folkard, ``Validation of the S and C components of the three-process model of alertness regulation,'' Sleep, vol. 18, no. 1, pp. 1–6, 1995. 


\bibitem{b12}
M. Sabet, R. A. Zoroofi, K. Sadeghniiat-Haghighi, and M. Sabbaghian, ``A new system for driver drowsiness and distraction detection,'' in Proceedings of The International Conference on E-Business and E-Government, pp. 1247–1251, May 2012. 

\bibitem{b13}
G. J. AL-Anizy, M. J. Nordin, and M. M. Razooq, ``Automatic driver drowsiness detection using haar algorithm and support vector machine techniques,'' Asian Journal of Applied Sciences, vol. 8, no. 2, pp. 149–157, 2015. 

\bibitem{b14}
L. Pauly and D. Sankar, ``Detection of drowsiness based on HOG features and SVM classifiers,'' in Proceedings of IEEE International Conference on Research in Computational Intelligence and Communication Networks, pp. 181–186, Nov. 2015. 

\bibitem{b15}
A. M. Bagci, R. Ansari, A. Khokhar, and E. Cetin, ``Eye tracking using
Markov models,''  in International Conference on Pattern Recognition, vol. 3, pp. 818–821, Aug. 2004. 

\bibitem{b16}
G. Pan, L. Sun, Z. Wu, and S. Lao, ``Eyeblink-based anti-spoofing in face recognition from a generic webcamera,''  in IEEE International Conference on Computer Vision, pp. 1–8, Oct. 2007. 

\bibitem{b17}
G. Pan, L. Sun, Z. Wu, and S. Lao, ``Eyeblink-based anti-spoofing in face recognition from a generic webcamera,'' in  IEEE International Conference on Computer Vision, pp. 1–8, Oct. 2007. 

\bibitem{b18}
K. Zhang, Z. Zhang, Z. Li, and Y. Qiao, ``Joint face detection and alignment using multitask cascaded convolutional networks,'' IEEE Signal Processing Letters, 23(10):1499–1503, 2016.

\bibitem{b19}
J. Hu, L. Shen, G. Sun,  ``Squeeze-and-excitation networks,'' in IEEE Conference on Computer Vision and Pattern Recognition, pp. 7132–7141, 2018.

\bibitem{b20}
R. Zhu, S. Zhang, X. Wang, L. Wen, H. Shi, L. Bo, and T. Mei, ``Scratchdet: Exploring to train single-shot object detectors from scratch,'' , in IEEE Conference on Computer Vision and Pattern Recognition, 2019.


\bibitem{b21}
T. H. Shih, and C. T. Hsu, ``MSTN: Multistage spatial-temporal network for driver drowsiness detection,'' in Asian Conference on Computer Vision. Springer, pp. 146–153, 2016.

\bibitem{b22}
T. H. VU,  A. DANG, J. C. WANG, ``A Deep Neural Network for Real-Time Driver Drowsiness Detection,'' IEICE Transaction on Information and Systems,  Vol.E102-D,  No. 12,  pp. 2637-2641, 2019.


\bibitem{b24}
X. Li, H. L. Chi, W. Zhang, and G. Q. Shen, ``Monitoring and Alerting of Crane Operator Fatigue Using Hybrid Deep Neural Network in the Prefabricated Products Assembly Process'' in International Symposium on Automation and Robotics in Construction, 2019. 

\end{thebibliography}
\end{document}